# Popular Ensemble Methods: An Empirical Study

**David Opitz**                                                                OPITZ@CS.UMT.EDU
*Department of Computer Science*
*University of Montana*
*Missoula, MT 59812 USA*

**Richard Maclin**                                                          RMACLIN@D.UMN.EDU
*Computer Science Department*
*University of Minnesota*
*Duluth, MN 55812 USA*

## Abstract

An ensemble consists of a set of individually trained classifiers (such as neural networks or decision trees) whose predictions are combined when classifying novel instances. Previous research has shown that an ensemble is often more accurate than any of the single classifiers in the ensemble. Bagging (Breiman, 1996c) and Boosting (Freund & Schapire, 1996; Schapire, 1990) are two relatively new but popular methods for producing ensembles. In this paper we evaluate these methods on 23 data sets using both neural networks and decision trees as our classification algorithm. Our results clearly indicate a number of conclusions. First, while Bagging is almost always more accurate than a single classifier, it is sometimes much less accurate than Boosting. On the other hand, Boosting can create ensembles that are less accurate than a single classifier – especially when using neural networks. Analysis indicates that the performance of the Boosting methods is dependent on the characteristics of the data set being examined. In fact, further results show that Boosting ensembles may overfit noisy data sets, thus decreasing its performance. Finally, consistent with previous studies, our work suggests that most of the gain in an ensemble's performance comes in the first few classifiers combined; however, relatively large gains can be seen up to 25 classifiers when Boosting decision trees.

## 1. Introduction

Many researchers have investigated the technique of combining the predictions of multiple classifiers to produce a single classifier (Breiman, 1996c; Clemen, 1989; Perrone, 1993; Wolpert, 1992). The resulting classifier (hereafter referred to as an *ensemble*) is generally more accurate than any of the individual classifiers making up the ensemble. Both theoretical (Hansen & Salamon, 1990; Krogh & Vedelsby, 1995) and empirical (Hashem, 1997; Opitz & Shavlik, 1996a, 1996b) research has demonstrated that a good ensemble is one where the individual classifiers in the ensemble are both accurate and make their errors on different parts of the input space. Two popular methods for creating accurate ensembles are Bagging (Breiman, 1996c) and Boosting (Freund & Schapire, 1996; Schapire, 1990). These methods rely on "resampling" techniques to obtain different training sets for each of the classifiers. In this paper we present a comprehensive evaluation of both Bagging and Boosting on 23 data sets using two basic classification methods: decision trees and neural networks.

Previous work has demonstrated that Bagging and Boosting are very effective for decision trees (Bauer & Kohavi, 1999; Drucker & Cortes, 1996; Breiman, 1996c, 1996b; Freund & Schapire, 1996; Quinlan, 1996); however, there has been little empirical testing with neural networks (especially with the new Boosting algorithm). Discussions with previous researchers reveal that many authors concentrated on decision trees due to their fast training speed and well-established default parameter settings. Neural networks present difficulties for testing both in terms of the significant processing time required and in selecting training parameters; however, we feel there are distinct advantages to including neural networks in our study. First, previous empirical studies have demonstrated that individual neural networks produce highly accurate classifiers that are sometimes more accurate than corresponding decision trees (Fisher & McKusick, 1989; Mooney, Shavlik, Towell, & Gove, 1989). Second, neural networks have been extensively applied across numerous domains (Arbib, 1995). Finally, by studying neural networks in addition to decision trees we can examine how Bagging and Boosting are influenced by the learning algorithm, giving further insight into the general characteristics of these approaches. Bauer and Kohavi (1999) also study Bagging and Boosting applied to two learning methods, in their case decision trees using a variant of C4.5 and naive-Bayes classifiers, but their study mainly concentrated on the decision tree results.

Our neural network and decision tree results led us to a number of interesting conclusions. The first is that a Bagging ensemble generally produces a classifier that is more accurate than a standard classifier. Thus one should feel comfortable always Bagging their decision trees or neural networks. For Boosting, however, we note more widely varying results. For a few data sets Boosting produced dramatic reductions in error (even compared to Bagging), but for other data sets it actually increases in error over a single classifier (particularly with neural networks). In further tests we examined the effects of noise and support Freund and Schapire's (1996) conjecture that Boosting's sensitivity to noise may be partly responsible for its occasional increase in error.

An alternate baseline approach we investigated was the creation of a simple neural network ensemble where each network used the full training set and differed only in its random initial weight settings. Our results indicate that this ensemble technique is surprisingly effective, often producing results as good as Bagging. Research by Ali and Pazzani (1996) demonstrated similar results using randomized decision tree algorithms.

Our results also show that the ensemble methods are generally consistent (in terms of their effect on accuracy) when applied either to neural networks or to decision trees; however, there is little inter-correlation between neural networks and decision trees except for the Boosting methods. This suggests that some of the increases produced by Boosting are dependent on the particular characteristics of the data set rather than on the component classifier. In further tests we demonstrate that Bagging is more resilient to noise than Boosting.

Finally, we investigated the question of how many component classifiers should be used in an ensemble. Consistent with previous research (Freund & Schapire, 1996; Quinlan, 1996), our results show that most of the reduction in error for ensemble methods occurs with the first few additional classifiers. With Boosting decision trees, however, relatively large gains may be seen up until about 25 classifiers.

This paper is organized as follows. In the next section we present an overview of classifier ensembles and discuss Bagging and Boosting in detail. Next we present an extensive empirical analysis of Bagging and Boosting. Following that we present future research and additional related work before concluding.

## 2. Classifier Ensembles

Figure 1 illustrates the basic framework for a classifier ensemble. In this example, neural networks are the basic classification method, though conceptually any classification method (e.g., decision trees) can be substituted in place of the networks. Each network in Figure 1's ensemble (network 1 through network $N$ in this case) is trained using the training instances for that network. Then, for each example, the predicted output of each of these networks ($o_i$ in Figure 1) is combined to produce the output of the ensemble ($\hat{o}$ in Figure 1). Many researchers (Alpaydin, 1993; Breiman, 1996c; Krogh & Vedelsby, 1995; Lincoln & Skrzypek, 1989) have demonstrated that an effective combining scheme is to simply average the predictions of the network.

Combining the output of several classifiers is useful only if there is disagreement among them. Obviously, combining several identical classifiers produces no gain. Hansen and Salamon (1990) proved that if the average error rate for an example is less than 50% and the component classifiers in the ensemble are independent in the production of their errors, the expected error for that example can be reduced to zero as the number of classifiers combined goes to infinity; however, such assumptions rarely hold in practice. Krogh and Vedelsby (1995) later proved that the ensemble error can be divided into a term measuring the average generalization error of each individual classifier and a term measuring the disagreement among the classifiers. What they formally showed was that an ideal ensemble consists of highly correct classifiers that disagree as much as possible. Opitz and Shavlik (1996a, 1996b) empirically verified that such ensembles generalize well.

As a result, methods for creating ensembles center around producing classifiers that disagree on their predictions. Generally, these methods focus on altering the training process in the hope that the resulting classifiers will produce different predictions. For example, neural network techniques that have been employed include methods for training with different topologies, different initial weights, different parameters, and training only on a portion of the training set (Alpaydin, 1993; Drucker, Cortes, Jackel, LeCun, & Vapnik, 1994; Hansen & Salamon, 1990; Maclin & Shavlik, 1995). In this paper we concentrate on two popular methods (Bagging and Boosting) that try to generate disagreement among the classifiers by altering the training set each classifier sees.

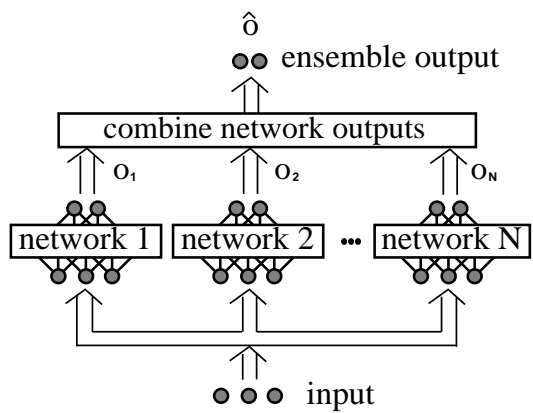

Figure 1: A classifier ensemble of neural networks.

## 2.1 Bagging Classifiers

Bagging (Breiman, 1996c) is a "bootstrap" (Efron & Tibshirani, 1993) ensemble method that creates individuals for its ensemble by training each classifier on a random redistribution of the training set. Each classifier's training set is generated by randomly drawing, with replacement, $N$ examples – where $N$ is the size of the original training set; many of the original examples may be repeated in the resulting training set while others may be left out. Each individual classifier in the ensemble is generated with a different random sampling of the training set.

Figure 2 gives a sample of how Bagging might work on a imaginary set of data. Since Bagging resamples the training set with replacement, some instance are represented multiple times while others are left out. So Bagging's training-set-1 might contain examples 3 and 7 twice, but does not contain either example 4 or 5. As a result, the classifier trained on training-set-1 might obtain a higher test-set error than the classifier using all of the data. In fact, all four of Bagging's component classifiers could result in higher test-set error; however, when combined, these four classifiers can (and often do) produce test-set error lower than that of the single classifier (the diversity among these classifiers generally compensates for the increase in error rate of any individual classifier).

Breiman (1996c) showed that Bagging is effective on "unstable" learning algorithms where small changes in the training set result in large changes in predictions. Breiman (1996c) claimed that neural networks and decision trees are examples of unstable learning algorithms. We study the effectiveness of Bagging on both these learning methods in this article.

## 2.2 Boosting Classifiers

Boosting (Freund & Schapire, 1996; Schapire, 1990) encompasses a family of methods. The focus of these methods is to produce a *series* of classifiers. The training set used for each member of the series is chosen based on the performance of the earlier classifier(s) in the series. In Boosting, examples that are incorrectly predicted by previous classifiers in the series are chosen more often than examples that were correctly predicted. Thus Boosting attempts to produce new classifiers that are better able to predict examples for which the current ensemble's performance is poor. (Note that in Bagging, the resampling of the training set is not dependent on the performance of the earlier classifiers.)

In this work we examine two new and powerful forms of Boosting: Arcing (Breiman, 1996b) and Ada-Boosting (Freund & Schapire, 1996). Like Bagging, Arcing chooses a training set of size $N$ for classifier $K + 1$ by probabilistically selecting (with replacement) examples from the original $N$ training examples. Unlike Bagging, however, the probability

| A sample of a single classifier on an imaginary set of data. | |
|---|---|
| | (Original) Training Set |
| Training-set-1: | 1, 2, 3, 4, 5, 6, 7, 8 |

| A sample of Bagging on the same data. | |
|---|---|
| | (Resampled) Training Set |
| Training-set-1: | 2, 7, 8, 3, 7, 6, 3, 1 |
| Training-set-2: | 7, 8, 5, 6, 4, 2, 7, 1 |
| Training-set-3: | 3, 6, 2, 7, 5, 6, 2, 2 |
| Training-set-4: | 4, 5, 1, 4, 6, 4, 3, 8 |

| A sample of Boosting on the same data. | |
|---|---|
| | (Resampled) Training Set |
| Training-set-1: | 2, 7, 8, 3, 7, 6, 3, 1 |
| Training-set-2: | 1, 4, 5, 4, 1, 5, 6, 4 |
| Training-set-3: | 7, 1, 5, 8, 1, 8, 1, 4 |
| Training-set-4: | 1, 1, 6, 1, 1, 3, 1, 5 |

Figure 2: Hypothetical runs of Bagging and Boosting. Assume there are eight training examples. Assume example 1 is an "outlier" and is hard for the component learning algorithm to classify correctly. With Bagging, each training set is an independent sample of the data; thus, some examples are missing and others occur multiple times. The Boosting training sets are also samples of the original data set, but the "hard" example (example 1) occurs more in later training sets since Boosting concentrates on correctly predicting it.

---

of selecting an example is not equal across the training set. This probability depends on how often that example was misclassified by the previous $K$ classifiers. Ada-Boosting can use the approach of (a) selecting a set of examples based on the probabilities of the examples, or (b) simply using all of the examples and weight the error of each example by the probability for that example (i.e., examples with higher probabilities have more effect on the error). This latter approach has the clear advantage that each example is incorporated (at least in part) in the training set. Furthermore, Friedman et al. (1998) have demonstrated that this form of Ada-Boosting can be viewed as a form of additive modeling for optimizing a logistic loss function. In this work, however, we have chosen to use the approach of subsampling the data to ensure a fair empirical comparison (in part due to the restarting reason discussed below).

Both Arcing and Ada-Boosting initially set the probability of picking each example to be $1/N$. These methods then recalculate these probabilities after each trained classifier is added to the ensemble. For Ada-Boosting, let $\epsilon_k$ be the sum of the probabilities of the misclassified instances for the currently trained classifier $C_k$. The probabilities for the next trial are generated by multiplying the probabilities of $C_k$'s incorrectly classified instances by the factor $\beta_k = (1 - \epsilon_k)/\epsilon_k$ and then renormalizing all probabilities so that their sum equals 1. Ada-Boosting combines the classifiers $C_1, \ldots, C_k$ using weighted voting where $C_k$ has weight $\log(\beta_k)$. These weights allow Ada-Boosting to discount the predictions of classifiers that are not very accurate on the overall problem. Friedman et al (1998) have also suggested an alternative mechanism that fits together the predictions of the classifiers as an additive model using a maximum likelihood criterion.

In this work, we use the revision described by Breiman (1996b) where we reset all the weights to be equal and restart if either $\epsilon_k$ is not less than 0.5 or $\epsilon_k$ becomes 0.[1] By resetting the weights we do not disadvantage the Ada-Boosting learner in those cases where it reaches these values of $\epsilon_k$; the Ada-Boosting learner always incorporates the same number of classifiers as other methods we tested. To make this feasible, we are forced to use the approach of selecting a data set probabilistically rather than weighting the examples, otherwise a deterministic method such as C4.5 would cycle and generate duplicate members of the ensemble. That is, resetting the weights to $1/N$ would cause the learner to repeat the decision tree learned as the first member of the ensemble, and this would lead to reweighting the data set the same as for the second member of the ensemble, and so on. Randomly selecting examples for the data set based on the example probabilities alleviates this problem.

Arcing-x4 (Breiman, 1996b) (which we will refer to simply as Arcing) started out as a simple mechanism for evaluating the effect of Boosting methods where the resulting classifiers were combined without weighting the votes. Arcing uses a simple mechanism for determining the probabilities of including examples in the training set. For the $i$th example in the training set, the value $m_i$ refers to the number of times that example was misclassified by the previous $K$ classifiers. The probability $p_i$ for selecting example $i$ to be part of classifier $K + 1$'s training set is defined as

$$p_i \;=\; \frac{1 + m_i{}^4}{\sum_{j=1}^{N}(1 + m_j{}^4)} \tag{1}$$

Breiman chose the value of the power (4) empirically after trying several different values (Breiman, 1996b). Although this mechanism does not have the weighted voting of Ada-Boosting it still produces accurate ensembles and is simple to implement; thus we include this method (along with Ada-Boosting) in our empirical evaluation.

Figure 2 shows a hypothetical run of Boosting. Note that the first training set would be the same as Bagging; however, later training sets accentuate examples that were misclassified by the earlier member of the ensembles. In this figure, example 1 is a "hard" example that previous classifiers tend to misclassify. With the second training set, example 1 occurs multiple times, as do examples 4 and 5 since they were left out of the first training set and, in this case, misclassified by the first learner. For the final training set, example 1 becomes the predominant example chosen (whereas no single example is accentuated with Bagging); thus, the overall test-set error for this classifier might become very high. Despite this, however, Boosting will probably obtain a lower error rate when it combines the output of these four classifiers since it focuses on correctly predicting previously misclassified examples and weights the predictions of the different classifiers based on their accuracy for the training set. But Boosting can also overfit in the presence of noise (as we empirically show in Section 3).

---

1. For those few cases where $\epsilon_k$ becomes 0 (less that 0.12% of our results) we simply use a large positive value, $log(\beta_k) = 3.0$, to weight these networks. For the more likely cases where $\epsilon_k$ is larger than 0.5 (approximately 5% of our results) we chose to weight the predictions by a very small positive value (0.001) rather than using a negative or 0 weight factor (this produced slightly better results than the alternate approaches in pilot studies).

## 2.3 The Bias plus Variance Decomposition

Recently, several authors (Breiman, 1996b; Friedman, 1996; Kohavi & Wolpert, 1996; Kong & Dietterich, 1995) have proposed theories for the effectiveness of Bagging and Boosting based on Geman et al.'s (1992) bias plus variance decomposition of classification error. In this decomposition we can view the expected error of a learning algorithm on a particular *target function* and *training set size* as having three components:

1. A *bias* term measuring how close the average classifier produced by the learning algorithm will be to the target function;

2. A *variance* term measuring how much each of the learning algorithm's guesses will vary with respect to each other (how often they disagree); and

3. A term measuring the minimum classification error associated with the Bayes optimal classifier for the target function (this term is sometimes referred to as the intrinsic target noise).

Using this framework it has been suggested (Breiman, 1996b) that both Bagging and Boosting reduce error by reducing the variance term. Freund and Schapire (1996) argue that Boosting also attempts to reduce the error in the bias term since it focuses on misclassified examples. Such a focus may cause the learner to produce an ensemble function that differs significantly from the single learning algorithm. In fact, Boosting may construct a function that is not even producible by its component learning algorithm (e.g., changing linear predictions into a classifier that contains non-linear predictions). It is this capability that makes Boosting an appropriate algorithm for combining the predictions of "weak" learning algorithms (i.e., algorithms that have a simple learning bias). In their recent paper, Bauer and Kohavi (1999) demonstrated that Boosting does indeed seem to reduce bias for certain real world problems. More surprisingly, they also showed that Bagging can also reduce the bias portion of the error, often for the same data sets for which Boosting reduces the bias.

Though the bias-variance decomposition is interesting, there are certain limitations to applying it to real-world data sets. To be able to estimate the bias, variance, and target noise for a particular problem, we need to know the actual function being learned. This is unavailable for most real-world problems. To deal with this problem Kohavi and Wolpert (1996) suggest holding out some of the data, the approach used by Bauer and Kohavi (1999) in their study. The main problem with this technique is that the training set size is greatly reduced in order to get good estimates of the bias and variance terms. We have chosen to strictly focus on generalization accuracy in our study, in part because Bauer and Kohavi's work has answered the question about whether Boosting and Bagging reduce the bias for real world problems (they both do), and because their experiments demonstrate that while this decomposition gives some insight into ensemble methods, it is only a small part of the equation. For different data sets they observe cases where Boosting and Bagging both decrease mostly the variance portion of the error, and other cases where Boosting and Bagging both reduce the bias and variance of the error. Their tests also seem to indicate that Boosting's generalization error increases on the domains where Boosting increases the variance portion of the error; but, it is difficult to determine what aspects of the data sets led to these results.

## 3. Results

This section describes our empirical study of Bagging, Ada-Boosting, and Arcing. Each of these three methods was tested with both decision trees and neural networks.

### 3.1 Data Sets

To evaluate the performance of Bagging and Boosting, we obtained a number of data sets from the University of Wisconsin Machine Learning repository as well as the UCI data set repository (Murphy & Aha, 1994). These data sets were hand selected such that they (a) came from real-world problems, (b) varied in characteristics, and (c) were deemed useful by previous researchers. Table 1 gives the characteristics of our data sets. The data sets chosen vary across a number of dimensions including: the type of the features in the data set (i.e., continuous, discrete, or a mix of the two); the number of output classes; and the number of examples in the data set. Table 1 also shows the architecture and training parameters used in our neural networks experiments.

### 3.2 Methodology

Results, unless otherwise noted, are averaged over five standard 10-fold cross validation experiments. For each 10-fold cross validation the data set is first partitioned into 10 equal-sized sets, then each set is in turn used as the test set while the classifier trains on the other nine sets. For each fold an ensemble of 25 classifiers is created. Cross validation folds were performed independently for each algorithm. We trained the neural networks using standard backpropagation learning (Rumelhart, Hinton, & Williams, 1986). Parameter settings for the neural networks include a learning rate of 0.15, a momentum term of 0.9, and weights are initialized randomly to be between -0.5 and 0.5. The number of hidden units and epochs used for training are given in the next section. We chose the number of hidden units based on the number of input and output units. This choice was based on the criteria of having at least one hidden unit per output, at least one hidden unit for every ten inputs, and five hidden units being a minimum. The number of epochs was based both on the number of examples and the number of parameters (i.e., topology) of the network. Specifically, we used 60 to 80 epochs for small problems involving fewer than 250 examples; 40 epochs for the mid-sized problems containing between 250 to 500 examples; and 20 to 40 epochs for larger problems. For the decision trees we used the C4.5 tool (Quinlan, 1993) and pruned trees (which empirically produce better performance) as suggested in Quinlan's work.

| Data Set | Cases | Class | Features | | Neural Network | | | |
|---|---|---|---|---|---|---|---|---|
| | | | Cont | Disc | Inputs | Outputs | Hiddens | Epochs |
| breast-cancer-w | 699 | 2 | 9 | - | 9 | 1 | 5 | 20 |
| credit-a | 690 | 2 | 6 | 9 | 47 | 1 | 10 | 35 |
| credit-g | 1000 | 2 | 7 | 13 | 63 | 1 | 10 | 30 |
| diabetes | 768 | 2 | 9 | - | 8 | 1 | 5 | 30 |
| glass | 214 | 6 | 9 | - | 9 | 6 | 10 | 80 |
| heart-cleveland | 303 | 2 | 8 | 5 | 13 | 1 | 5 | 40 |
| hepatitis | 155 | 2 | 6 | 13 | 32 | 1 | 10 | 60 |
| house-votes-84 | 435 | 2 | - | 16 | 16 | 1 | 5 | 40 |
| hypo | 3772 | 5 | 7 | 22 | 55 | 5 | 15 | 40 |
| ionosphere | 351 | 2 | 34 | - | 34 | 1 | 10 | 40 |
| iris | 159 | 3 | 4 | - | 4 | 3 | 5 | 80 |
| kr-vs-kp | 3196 | 2 | - | 36 | 74 | 1 | 15 | 20 |
| labor | 57 | 2 | 8 | 8 | 29 | 1 | 10 | 80 |
| letter | 20000 | 26 | 16 | - | 16 | 26 | 40 | 30 |
| promoters-936 | 936 | 2 | - | 57 | 228 | 1 | 20 | 30 |
| ribosome-bind | 1877 | 2 | - | 49 | 196 | 1 | 20 | 35 |
| satellite | 6435 | 6 | 36 | - | 36 | 6 | 15 | 30 |
| segmentation | 2310 | 7 | 19 | - | 19 | 7 | 15 | 20 |
| sick | 3772 | 2 | 7 | 22 | 55 | 1 | 10 | 40 |
| sonar | 208 | 2 | 60 | - | 60 | 1 | 10 | 60 |
| soybean | 683 | 19 | - | 35 | 134 | 19 | 25 | 40 |
| splice | 3190 | 3 | - | 60 | 240 | 2 | 25 | 30 |
| vehicle | 846 | 4 | 18 | - | 18 | 4 | 10 | 40 |

Table 1: Summary of the data sets used in this paper. Shown are the number of examples in the data set; the number of output classes; the number of continuous and discrete input features; the number of input, output, and hidden units used in the neural networks tested; and how many epochs each neural network was trained.

### 3.3 Data Set Error Rates

Table 2 shows test-set error rates for the data sets described in Table 1 for five neural network methods and four decision tree methods. (In Tables 4 and 5 we show these error rates as well as the standard deviation for each of these values.) Along with the test-set errors for Bagging, Arcing, and Ada-boosting, we include the test-set error rate for a single neural-network and a single decision-tree classifier. We also report results for a simple (baseline) neural-network ensemble approach – creating an ensemble of networks where each network varies only by randomly initializing the weights of the network. We include these results in certain comparisons to demonstrate their similarity to Bagging. One obvious conclusion drawn from the results is that each ensemble method appears to reduce the error rate for almost all of the data sets, and in many cases this reduction is large. In fact, the two-tailed sign test indicates that every ensemble method is significantly better than its single component classifier at the 95% confidence level; however, none of the ensemble methods are significantly better than any other ensemble approach at the 95% confidence level.

| Data Set | Neural Network | | | | | C4.5 | | | |
|---|---|---|---|---|---|---|---|---|---|
| | | | | Boosting | | | | Boosting | |
| | Stan | Simp | Bag | Arc | Ada | Stan | Bag | Arc | Ada |
| breast-cancer-w | 3.4 | 3.5 | 3.4 | 3.8 | 4.0 | 5.0 | 3.7 | 3.5 | 3.5 |
| credit-a | 14.8 | 13.7 | 13.8 | 15.8 | 15.7 | 14.9 | 13.4 | 14.0 | 13.7 |
| credit-g | 27.9 | 24.7 | 24.2 | 25.2 | 25.3 | 29.6 | 25.2 | 25.9 | 26.7 |
| diabetes | 23.9 | 23.0 | 22.8 | 24.4 | 23.3 | 27.8 | 24.4 | 26.0 | 25.7 |
| glass | 38.6 | 35.2 | 33.1 | 32.0 | 31.1 | 31.3 | 25.8 | 25.5 | 23.3 |
| heart-cleveland | 18.6 | 17.4 | 17.0 | 20.7 | 21.1 | 24.3 | 19.5 | 21.5 | 20.8 |
| hepatitis | 20.1 | 19.5 | 17.8 | 19.0 | 19.7 | 21.2 | 17.3 | 16.9 | 17.2 |
| house-votes-84 | 4.9 | 4.8 | 4.1 | 5.1 | 5.3 | 3.6 | 3.6 | 5.0 | 4.8 |
| hypo | 6.4 | 6.2 | 6.2 | 6.2 | 6.2 | 0.5 | 0.4 | 0.4 | 0.4 |
| ionosphere | 9.7 | 7.5 | 9.2 | 7.6 | 8.3 | 8.1 | 6.4 | 6.0 | 6.1 |
| iris | 4.3 | 3.9 | 4.0 | 3.7 | 3.9 | 5.2 | 4.9 | 5.1 | 5.6 |
| kr-vs-kp | 2.3 | 0.8 | 0.8 | 0.4 | 0.3 | 0.6 | 0.6 | 0.3 | 0.4 |
| labor | 6.1 | 3.2 | 4.2 | 3.2 | 3.2 | 16.5 | 13.7 | 13.0 | 11.6 |
| letter | 18.0 | 12.8 | 10.5 | 5.7 | 4.6 | 14.0 | 7.0 | 4.1 | 3.9 |
| promoters-936 | 5.3 | 4.8 | 4.0 | 4.5 | 4.6 | 12.8 | 10.6 | 6.8 | 6.4 |
| ribosome-bind | 9.3 | 8.5 | 8.4 | 8.1 | 8.2 | 11.2 | 10.2 | 9.3 | 9.6 |
| satellite | 13.0 | 10.9 | 10.6 | 9.9 | 10.0 | 13.8 | 9.9 | 8.6 | 8.4 |
| segmentation | 6.6 | 5.3 | 5.4 | 3.5 | 3.3 | 3.7 | 3.0 | 1.7 | 1.5 |
| sick | 5.9 | 5.7 | 5.7 | 4.7 | 4.5 | 1.3 | 1.2 | 1.1 | 1.0 |
| sonar | 16.6 | 15.9 | 16.8 | 12.9 | 13.0 | 29.7 | 25.3 | 21.5 | 21.7 |
| soybean | 9.2 | 6.7 | 6.9 | 6.7 | 6.3 | 8.0 | 7.9 | 7.2 | 6.7 |
| splice | 4.7 | 4.0 | 3.9 | 4.0 | 4.2 | 5.9 | 5.4 | 5.1 | 5.3 |
| vehicle | 24.9 | 21.2 | 20.7 | 19.1 | 19.7 | 29.4 | 27.1 | 22.5 | 22.9 |

Table 2:   Test set error rates for the data sets using (1) a single neural network classifier; (2) an ensemble where each individual network is trained using the original training set and thus only differs from the other networks in the ensemble by its random initial weights; (3) an ensemble where the networks are trained using randomly resampled training sets (Bagging); an ensemble where the networks are trained using weighted resampled training sets (Boosting) where the resampling is based on the (4) Arcing method and (5) Ada method; (6) a single decision tree classifier; (7) a Bagging ensemble of decision trees; and (8) Arcing and (9) Ada Boosting ensembles of decision trees.

To better analyze Table 2's results, Figures 3 and 4 plot the percentage reduction in error for the Ada-Boosting, Arcing, and Bagging method as a function of the original error rate. Examining these figures we note that many of the gains produced by the ensemble methods are much larger than the standard deviation values. In terms of comparisons of different methods, it is apparent from both figures that the Boosting methods (Ada-

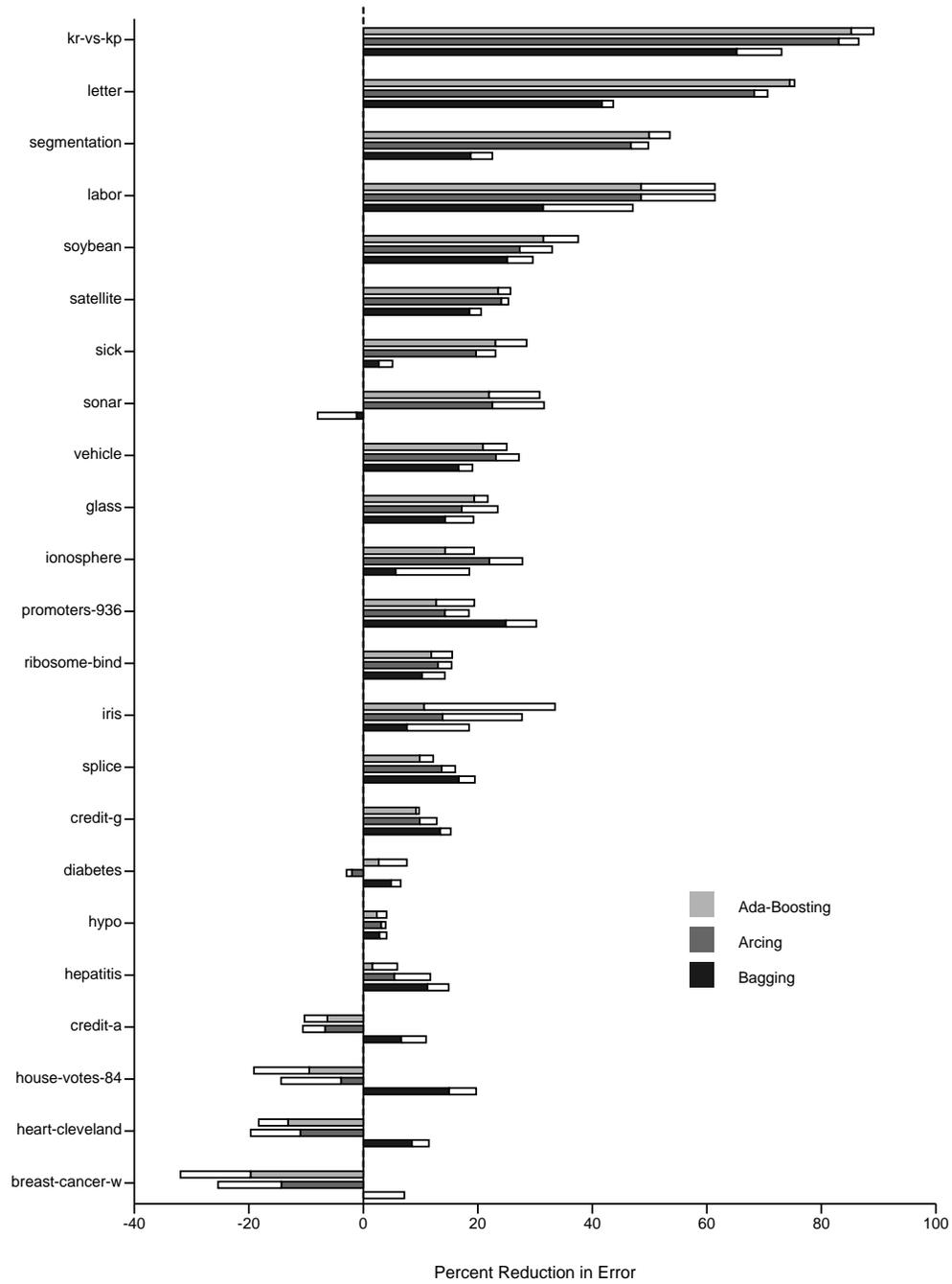

Figure 3: Reduction in error for Ada-Boosting, Arcing, and Bagging neural network ensembles as a percentage of the original error rate (i.e., a reduction from an error rate of 2.5% to 1.25% would be a 50% reduction in error rate, just as a reduction from 10.0% to 5.0% would also be a 50% reduction). Also shown (white portion of each bar) is one standard deviation for these results. The standard deviation is shown as an addition to the error reduction.

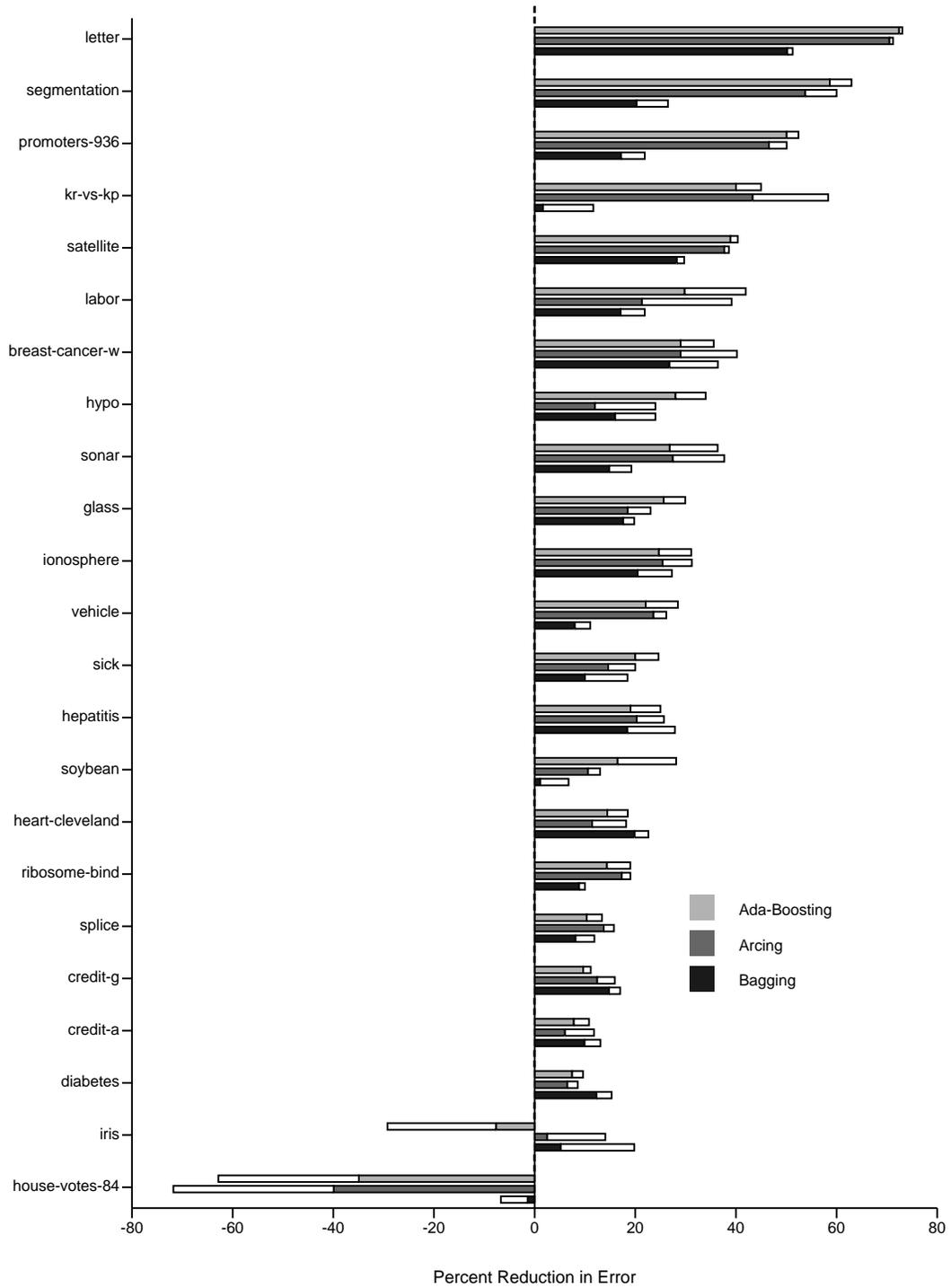

Figure 4: Reduction in error for Ada-Boosting, Arcing, and Bagging decision tree ensembles as a percentage of the original error rate. Also shown (white portion of each bar) is one standard deviation for these results.

Boosting and Arcing) are similar in their results, both for neural networks and decision trees. Furthermore, the Ada-Boosting and Arcing methods produce some of the largest reductions in error. On the other hand, while the Bagging method consistently produces reductions in error for almost all of the cases, with neural networks the Boosting methods can sometimes result in an increase in error.

Looking at the ordering of the data sets in the two figures (the results are sorted by the percentage of reduction using the Ada-Boosting method), we note that the data sets for which the ensemble methods seem to work well are somewhat consistent across both neural networks and decision trees. For the few domains which see increases in error, it is difficult to reach strong conclusions since the ensemble methods seem to do well for a large number of domains. One domain on which the Boosting methods do uniformly poorly is the house-votes-84 domain. As we discuss later, there may noise in this domain's examples that causes the Boosting methods significant problems.

### 3.4 Ensemble Size

Early work (Hansen & Salamon, 1990) on ensembles suggested that ensembles with as few as ten members were adequate to sufficiently reduce test-set error. While this claim may be true for the earlier proposed ensembles, the Boosting literature (Schapire, Freund, Bartlett, & Lee, 1997) has recently suggested (based on a few data sets with decision trees) that it is possible to further reduce test-set error even after ten members have been added to an ensemble (and they note that this result also applies to Bagging). In this section, we perform additional experiments to further investigate the appropriate size of an ensemble. Figure 5 shows the composite error rate over all of our data sets for neural network and decision tree ensembles using up to 100 classifiers. Our experiments indicate that most of the methods produce similarly shaped curves. As expected, much of the reduction in error due to adding classifiers to an ensemble comes with the first few classifiers; however, there is some variation with respect to where the error reduction finally asymptotes.

For both Bagging and Boosting applied to neural networks, much of the reduction in error appears to have occurred after ten to fifteen classifiers. A similar conclusion can be reached for Bagging and decision trees, which is consistent with Breiman (1996a). But Ada-boosting and Arcing continue to measurably improve their test-set error until around 25 classifiers for decision trees. At 25 classifiers the error reduction for both methods appears to have nearly asymptoted to a plateau. Therefore, the results reported in this paper are of an ensemble size of 25 (i.e., a sufficient yet manageable size for qualitative analysis). It was traditionally believed (Freund & Schapire, 1996) that small reductions in test-set error may continue indefinitely for boosting; however, Grove and Schuurmans (1998) demonstrate that Ada-boosting can indeed begin to overfit with *very* large ensemble sizes (10,000 or more members).

### 3.5 Correlation Among Methods

As suggested above, it appears that the performance of many of the ensemble methods are highly correlated with one another. To help identify these consistencies, Table 3 presents the correlation coefficients of the performance of all seven ensemble methods. For each data set, performance is measured as the ensemble error rate divided by the single-classifier error rate. Thus a high correlation (i.e., one near 1.0) suggests that two methods are consistent in the domains in which they have the greatest impact on test-set error reduction.

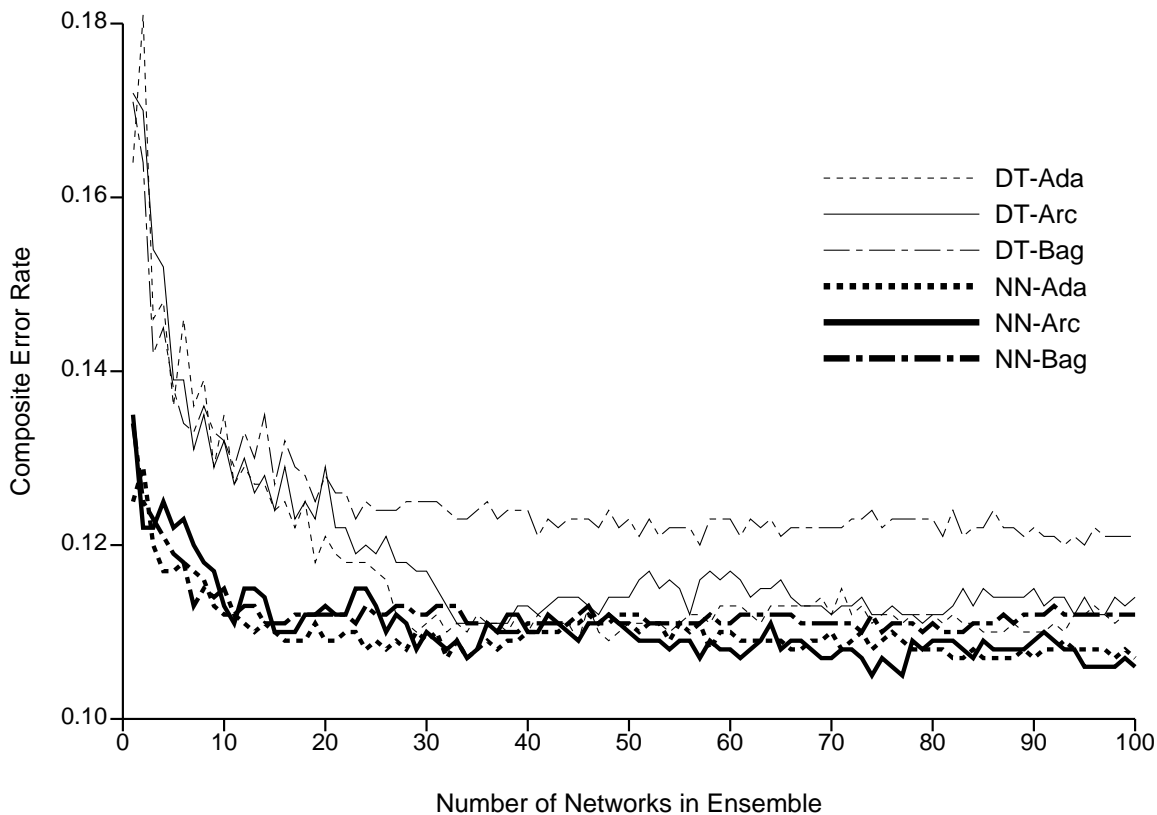

Figure 5: Average test-set error over all 23 data sets used in our studies for ensembles incorporating from one to 100 decision trees or neural networks. The error rate graphed is simply the average of the error rates of the 23 data sets. The alternative of averaging the error over all data points (i.e., weighting a data set's error rate by its sample size) produces similarly shaped curves.

Table 3 provides numerous interesting insights. The first is that the neural-network ensemble methods are strongly correlated with one another and the decision-tree ensemble methods are strongly correlated with one another; however, there is less correlation between any neural-network ensemble method and any decision-tree ensemble method. Not surprisingly, Ada-boosting and Arcing are strongly correlated, even across different component learning algorithms. This suggests that Boosting's effectiveness depends more on the data set than whether the component learning algorithm is a neural network or decision tree. Bagging on the other hand, is not correlated across component learning algorithms. These results are consistent with our later claim that while Boosting is a powerful ensemble method, it is more susceptible to a noisy data set than Bagging.

| | Neural Network | | | | Decision Tree | | |
|---|---|---|---|---|---|---|---|
| | Simple | Bagging | Arcing | Ada | Bagging | Arcing | Ada |
| Simple-NN | 1.00 | 0.88 | 0.87 | 0.85 | -0.10 | 0.38 | 0.37 |
| Bagging-NN | 0.88 | 1.00 | 0.78 | 0.78 | -0.11 | 0.35 | 0.35 |
| Arcing-NN | 0.87 | 0.78 | 1.00 | 0.99 | 0.14 | 0.61 | 0.60 |
| Ada-NN | 0.85 | 0.78 | 0.99 | 1.00 | 0.17 | 0.62 | 0.63 |
| Bagging-DT | -0.10 | -0.11 | 0.14 | 0.17 | 1.00 | 0.68 | 0.69 |
| Arcing-DT | 0.38 | 0.35 | 0.61 | 0.62 | 0.68 | 1.00 | 0.96 |
| Ada-DT | 0.37 | 0.35 | 0.60 | 0.63 | 0.69 | 0.96 | 1.00 |

Table 3: Performance correlation coefficients across ensemble learning methods. Performance is measured by the ratio of the ensemble method's test-set error divided by the single component classifier's test-set error.

### 3.6 Bagging versus Simple network ensembles

Figure 6 shows the Bagging and Simple network ensemble results from Table 2. These results indicate that often a Simple Ensemble approach will produce results that are as accurate as Bagging (correlation results from Table 3 also support this statement). This suggests that any mechanism which causes a learning method to produce some randomness in the formation of its classifiers can be used to form accurate ensembles, and indeed, Ali and Pazzani (1996) have demonstrated similar results for randomized decision trees.

### 3.7 Neural Networks versus Decision Trees

Another interesting question is how effective the different methods are for neural networks and decision trees. Figures 7, 8, and 9 compare the error rates and reduction in error values for Ada-Boosting, Arcing, and Bagging respectively. Note that we graph error rate rather than percent reduction in error rate because the baseline for each method (decision trees for Ada-Boosting on decision trees versus neural networks for Ada-Boosting on neural networks) may partially explain the differences in percent reduction. For example, in the promoters-936 problem using Ada-Boosting, the much larger reduction in error for the decision tree approach may be due to the fact that decision trees do not seem to be as effective for this problem, and Ada-Boosting therefore produces a larger percent reduction in the error for decision trees.

The results show that in many cases if a single decision tree had lower (or higher) error than a single neural network on a data set, then the decision-tree ensemble methods also had lower (or higher) error than their network counterpart. The exceptions to this rule generally happened on the same data set for all three ensemble methods (e.g., hepatitis, soybean, satellite, credit-a, and heart-cleveland). These results suggest that (a) the performance of the ensemble methods is dependent on both the data set and classifier method, and (b) ensembles can, at least in some cases, overcome the inductive bias of its component learning algorithm.

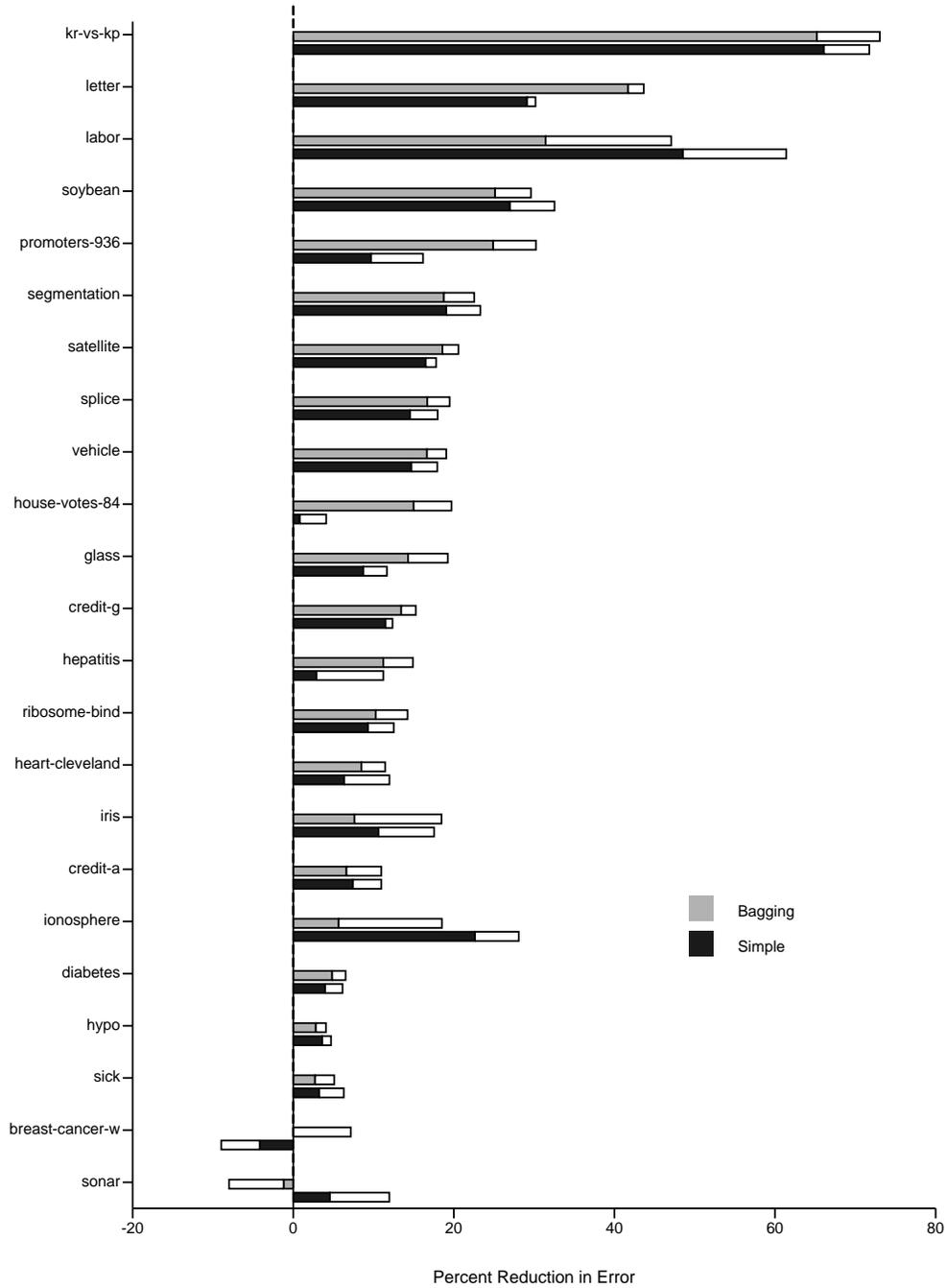

Figure 6: Reduction in error for Bagging and Simple neural network ensembles as a percentage of the original error rate. Also shown (white portion of each bar) is one standard deviation for these results.

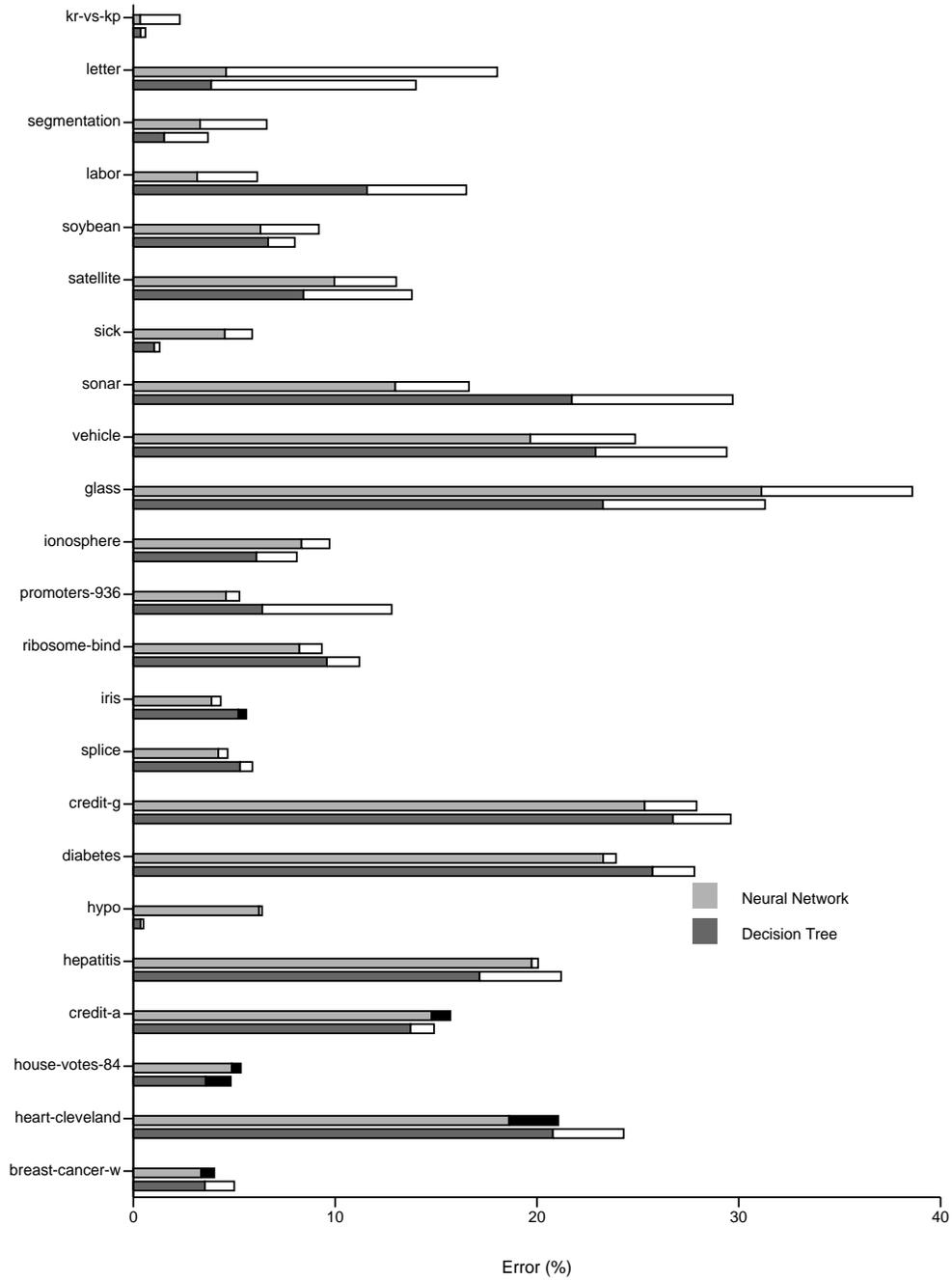

Figure 7: Error rates for Ada-Boosting ensembles. The white portion shows the reduction in error of Ada-Boosting compared to a single classifier while increases in error are shown in black. The data sets are sorted by the ratio of reduction in ensemble error to overall error for neural networks.

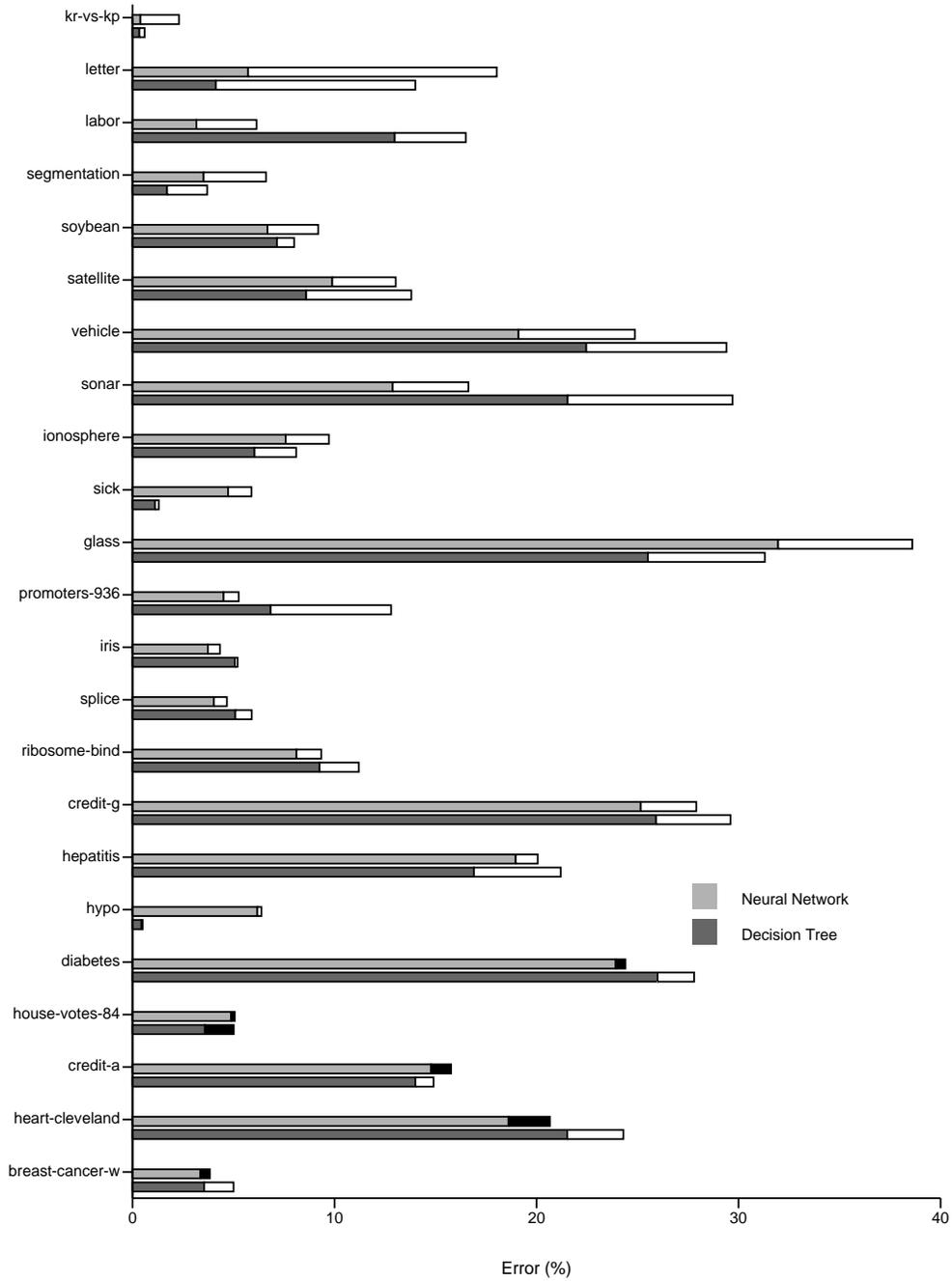

Figure 8: Error rates for Arcing ensembles. The white portion shows the reduction in error of Arcing compared to a single classifier while increases in error are shown in black. The data sets are sorted by the ratio of reduction in ensemble error to overall error for neural networks.

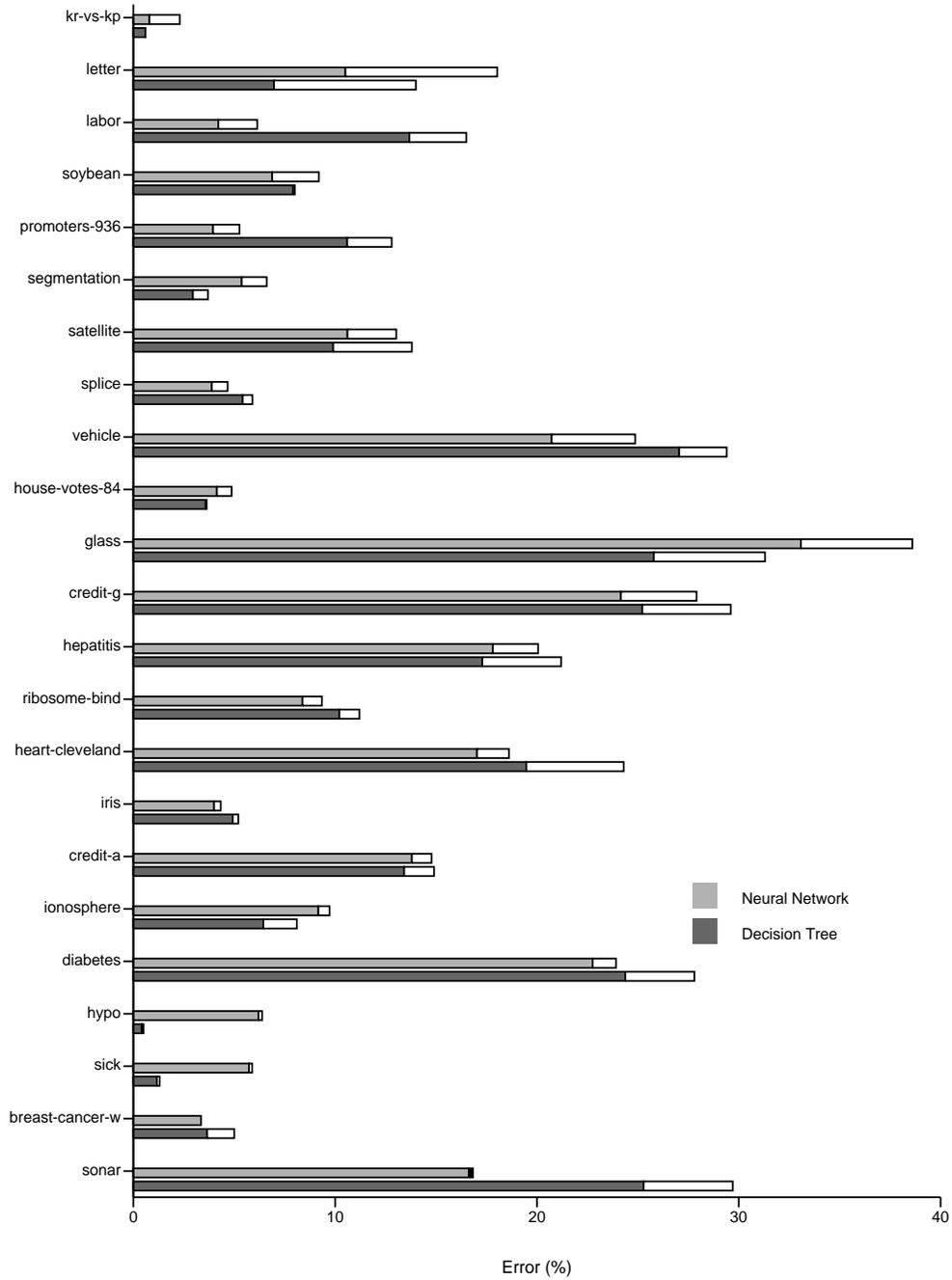

Figure 9: Error rates for Bagging ensembles. The white portion shows the reduction in error of Bagging compared to a single classifier while increases in error are shown in black. The data sets are sorted by the ratio of reduction in ensemble error to overall error for neural networks.

### 3.8 Boosting and Noise

Freund and Shapire (1996) suggested that the sometimes poor performance of Boosting results from overfitting the training set since later training sets may be over-emphasizing examples that are noise (thus creating extremely poor classifiers). This argument seems especially pertinent to Boosting for two reasons. The first and most obvious reason is that their method for updating the probabilities may be over-emphasizing noisy examples. The second reason is that the classifiers are combined using weighted voting. Previous work (Sollich & Krogh, 1996) has shown that optimizing the combining weights can lead to overfitting while an unweighted voting scheme is generally resilient to overfitting. Friedman et al. (1998) hypothesize that Boosting methods, as additive models, may see increases in error in those situations where the bias of the base classifier is appropriate for the problem being learned. We test this hypothesis in our second set of results presented in this section.

To evaluate the hypothesis that Boosting may be prone to overfitting we performed a set of experiments using the four ensemble neural network methods. We introduced 5%, 10%, 20%, and 30% noise[2] into four different data sets. At each level we created five different noisy data sets, performed a 10-fold cross validation on each, then averaged over the five results. In Figure 10 we show the reduction in error rate for each of the ensemble methods compared to using a single neural network classifier. These results demonstrate that as the noise level grows, the efficacy of the Simple and Bagging ensembles generally increases while the Arcing and Ada-Boosting ensembles gains in performance are much smaller (or may actually decrease). Note that this effect is more extreme for Ada-Boosting which supports our hypothesis that Ada-Boosting is more affected by noise. This suggests that Boosting's poor performance for certain data sets may be partially explained by overfitting noise.

To further demonstrate the effect of noise on Boosting we created several sets of artificial data specifically designed to mislead Boosting methods. For each data set we created a simple hyperplane concept based on a set of the features (and also included some irrelevant features). A set of random points were then generated and labeled based on which side of the hyperplane they fell. Then a certain percentage of the points on one side of the hyperplane were mislabeled as being part of the other class. For the experiments shown below we generated five data sets where the concept was based on two linear features, had four irrelevant features, and 20% of the data was mislabeled. We trained five ensembles of neural networks (perceptrons) for each data set and averaged the ensembles' predictions. Thus these experiments involve learning in situations where the original bias of the learner (a single hyperplane produced by a perceptron) is appropriate for the problem, and as Friedman et al. (1998) suggest, using an additive model may harm performance. Figure 11 shows the resulting error rates for Ada-Boosting, Arcing, and Bagging by the number of networks being combined in the ensemble. These results indicate clearly that in cases where there is noise Bagging's error rate will not increase as the ensemble size increases whereas the error rate of the Boosting methods may indeed increase as ensemble size increases.

---

2. X% noise indicates that each feature of the training examples, both input and output features, had X% chance of being randomly perturbed to another feature value for that feature (for continuous features, the set of possible other values was chosen by examining all of the training examples).

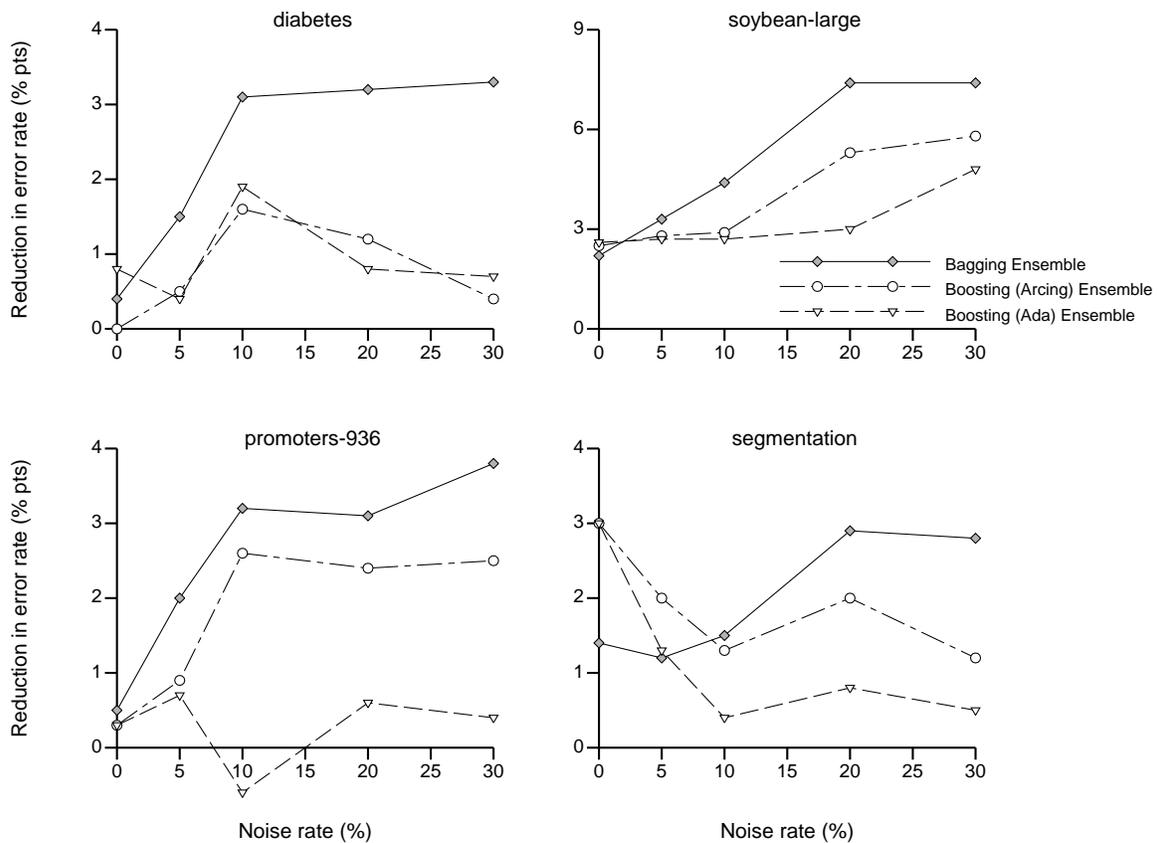

Figure 10: Simple, Bagging, and Boosting (Arcing and Ada) neural network ensemble reduction in error as compared to using a single neural network. Graphed is the percentage *point* reduction in error (e.g., for 5% noise in the segmentation data set, if the single network method had an error rate of 15.9% and the Bagging method had an error rate of 14.7%, then this is graphed as a 1.2 percentage point reduction in the error rate).

Additional tests (not shown here) show that Ada-Boosting's error rate becomes worse when restarting is not employed.

This conclusion dovetails nicely with Schapire et al.'s (1997) recent discussion where they note that the effectiveness of a voting method can be measured by examining the *margins* of the examples. (The margin is the difference between the number of correct and incorrect votes for an example.) In a simple resampling method such as Bagging, each resulting classifier focuses on increasing the margin for as many of the examples as possible. But in a Boosting method, later classifiers focus on increasing the margins for examples with poor current margins. As Schapire et al. (1997) note, this is a very effective strategy if the overall accuracy of the resulting classifier does not drop significantly. For a problem with noise, focusing on misclassified examples may cause a classifier to focus on boosting the margins of (noisy) examples that would in fact be misleading in overall classification.

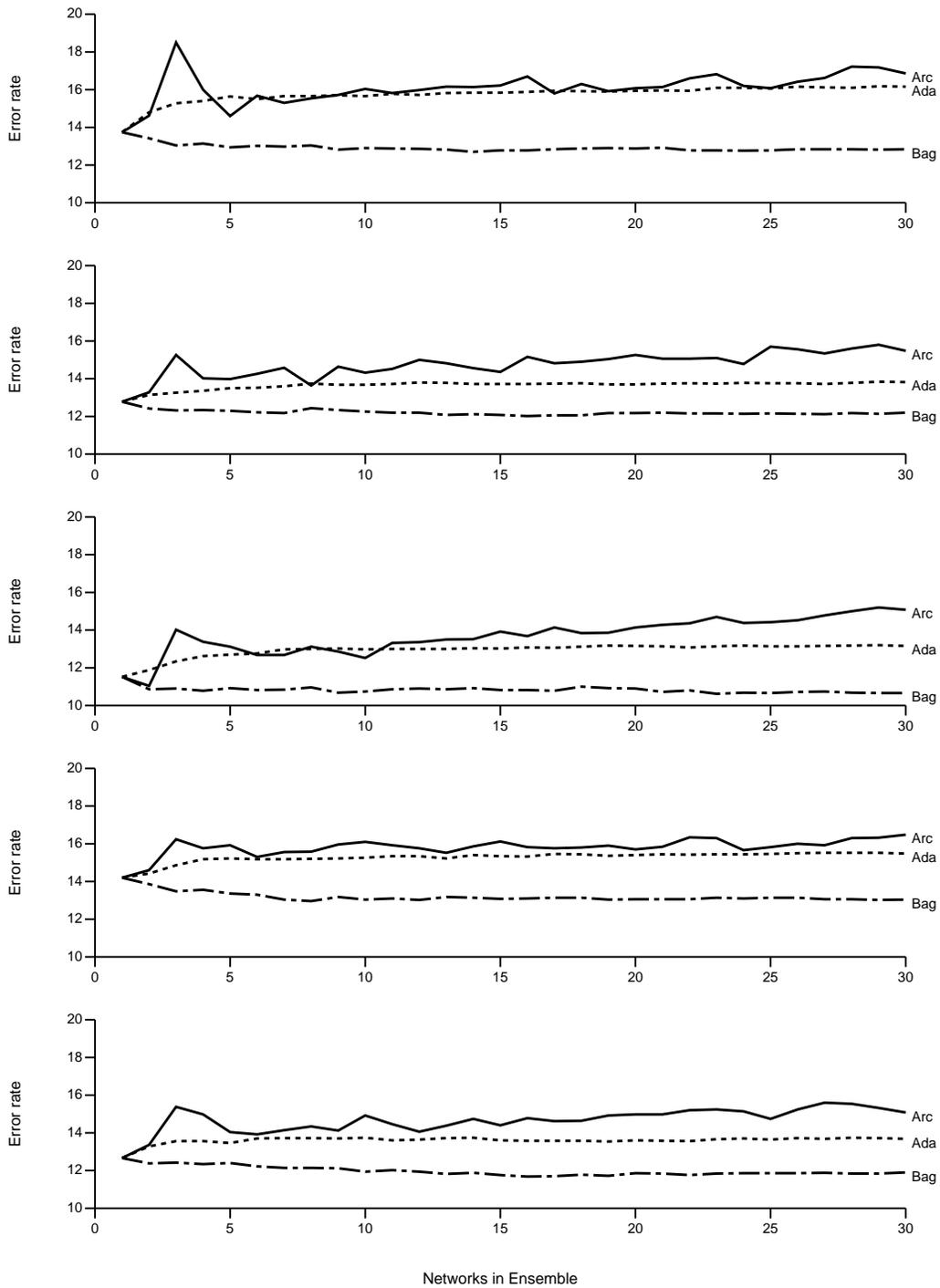

Figure 11: Error rates by the size of ensemble for Ada-Boosting, Arcing, and Bagging ensembles for five different artificial data sets containing one-sided noise (see text for description).

## 4. Future Work

One interesting question we plan to investigate is how effective a single classifier approach might be if it was allowed to use the time it takes the ensemble method to train multiple classifiers to explore its concept space. For example, a neural network approach could perform pilot studies using the training set to select appropriate values of parameters such as hidden units, learning rate, etc.

We plan to compare Bagging and Boosting methods to other methods introduced recently. In particular we intend to examine the use of Stacking (Wolpert, 1992) as a method of *training* a combining function, so as to avoid the effect of having to weight classifiers. We also plan to compare Bagging and Boosting to other methods such as Opitz and Shavlik's (1996b) approach to creating an ensemble. This approach uses genetic search to find classifiers that are accurate and differ in their predictions.

Finally, since the Boosting methods are extremely successful in many domains, we plan to investigate novel approaches that will retain the benefits of Boosting. The goal will be to create a learner where you can essentially push a start button and let it run. To do this we would try to preserve the benefits of Boosting while preventing overfitting on noisy data sets. One possible approach would be to use a holdout training set (a tuning set) to evaluate the performance of the Boosting ensemble to determine when the accuracy is no longer increasing. Another approach would be to use pilot studies to determine an "optimal" number of classifiers to use in an ensemble.

## 5. Additional Related Work

As mentioned before, the idea of using an ensemble of classifiers rather than the single best classifier has been proposed by several people. In Section 2, we present a framework for these systems, some theories of what makes an effective ensemble, an extensive covering of the Bagging and Boosting algorithms, and a discussion on the bias plus variance decomposition. Section 3 referred to empirical studies similar to ours; these methods differ from ours in that they were limited to decision trees, generally with fewer data sets. We cover additional related work in this section.

Lincoln and Skrzypek (1989), Mani (1991) and the forecasting literature (Clemen, 1989; Granger, 1989) indicate that a simple averaging of the predictors generates a very good composite model; however, many later researchers (Alpaydin, 1993; Asker & Maclin, 1997a, 1997b; Breiman, 1996c; Hashem, 1997; Maclin, 1998; Perrone, 1992; Wolpert, 1992; Zhang, Mesirov, & Waltz, 1992) have further improved generalization with voting schemes that are complex combinations of each predictor's output. One must be careful in this case, since optimizing the combining weights can easily lead to the problem of overfitting which simple averaging seems to avoid (Sollich & Krogh, 1996).

Most approaches only *indirectly* try to generate highly correct classifiers that disagree as much as possible. These methods try to create diverse classifiers by training classifiers with dissimilar learning parameters (Alpaydin, 1993), different classifier architectures (Hashem, 1997), various initial neural-network weight settings (Maclin & Opitz, 1997; Maclin & Shavlik, 1995), or separate partitions of the training set (Breiman, 1996a; Krogh & Vedelsby, 1995). Boosting on the other hand is active in trying to generate highly correct networks since it accentuates examples currently classified incorrectly by previous members of the ensemble.

ADDEMUP (Opitz & Shavlik, 1996a, 1996b) is another example of an approach that directly tries to create a diverse ensemble. ADDEMUP uses genetic algorithms to search explicitly for a highly diverse set of accurate trained networks. ADDEMUP works by first creating an initial population, then uses genetic operators to create new networks continually, keeping the set of networks that are highly accurate while disagreeing with each other as much as possible. ADDEMUP is also effective at incorporating prior knowledge, if available, to improve the quality of its ensemble.

An alternate approach to the ensemble framework is to train individual networks on a subtask, and to then combine these predictions with a "gating" function that depends on the input. Jacobs et al.'s (1991) adaptive mixtures of local experts, Baxt's (1992) method for identifying myocardial infarction, and Nowlan and Sejnowski's (1992) visual model all train networks to learn specific subtasks. The key idea of these techniques is that a decomposition of the problem into specific subtasks might lead to more efficient representations and training (Hampshire & Waibel, 1989).

Once a problem is broken into subtasks, the resulting solutions need to be combined. Jacobs et al. (1991) propose having the gating function be a network that *learns* how to allocate examples to the experts. Thus the gating network allocates each example to one or more experts, and the backpropagated errors and resulting weight changes are then restricted to these networks (and the gating function). Tresp and Taniguchi (1995) propose a method for determining the gating function *after* the problem has been decomposed and the experts trained. Their gating function is an input-dependent, linear-weighting function that is determined by a combination of the networks' diversity on the current input with the likelihood that these networks have seen data "near" that input.

Although the mixtures of experts and ensemble paradigms seem very similar, they are in fact quite distinct from a statistical point of view. The mixtures-of-experts model makes the assumption that a single expert is responsible for each example. In this case, each expert is a model of a region of the input space, and the job of the gating function is to decide from which model the data point originates. Since each network in the ensemble approach learns the whole task rather than just some subtask and thus makes no such mutual exclusivity assumption, ensembles are appropriate when no one model is highly likely to be correct for any one point in the input space.

## 6. Conclusions

This paper presents a comprehensive empirical evaluation of Bagging and Boosting for neural networks and decision trees. Our results demonstrate that a Bagging ensemble nearly always outperforms a single classifier. Our results also show that a Boosting ensemble can greatly outperform both Bagging and a single classifier. However, for some data sets Boosting may show zero gain or even a decrease in performance from a single classifier. Further tests indicate that Boosting may suffer from overfitting in the presence of noise which may explain some of the decreases in performance for Boosting. We also found that a simple ensemble approach of using neural networks that differ only in their random initial weight settings performed surprisingly well, often doing as well as the Bagging.

Analysis of our results suggests that the performance of both Boosting methods (Ada-Boosting and Arcing) is at least partly dependent on the data set being examined, where Bagging shows much less correlation. The strong correlations for Boosting may be partially explained by its sensitivity to noise, a claim supported by additional tests. Finally, we show that much of the performance enhancement for an ensemble comes with the first few classifiers combined, but that Boosting decision trees may continue to further improve with larger ensemble sizes.

In conclusion, as a general technique for decision trees and neural networks, Bagging is probably appropriate for most problems, but when appropriate, Boosting (either Arcing or Ada) may produce larger gains in accuracy.

## Acknowledgments

This research was partially supported by University of Minnesota Grants-in-Aid to both authors. Dave Opitz was also supported by National Science Foundation grant IRI-9734419, the Montana DOE/EPSCoR Petroleum Reservoir Characterization Project, a MONTS grant supported by the University of Montana, and a Montana Science Technology Alliance grant. This is an extended version of a paper published in the *Fourteenth National Conference on Artificial Intelligence*.

# Appendix

Tables 4 and 5 show the complete results for the first set of experiments used in this paper.

| Data Set | Single | | | Simple | | Bagging | | Arcing | | Boosting | |
|---|---|---|---|---|---|---|---|---|---|---|---|
| | Err | SD | Best | Err | SD | Err | SD | Err | SD | Err | SD |
| breast-cancer-w | 3.4 | 0.3 | 2.9 | 3.5 | 0.2 | 3.4 | 0.2 | 3.8 | 0.4 | 4.0 | 0.4 |
| credit-a | 14.8 | 0.7 | 13.6 | 13.7 | 0.5 | 13.8 | 0.6 | 15.8 | 0.6 | 15.7 | 0.6 |
| credit-g | 27.9 | 0.8 | 26.2 | 24.7 | 0.2 | 24.2 | 0.5 | 25.2 | 0.8 | 25.3 | 0.1 |
| diabetes | 23.9 | 0.9 | 22.6 | 23.0 | 0.5 | 22.8 | 0.4 | 24.4 | 0.2 | 23.3 | 1.2 |
| glass | 38.6 | 1.5 | 36.9 | 35.2 | 1.1 | 33.1 | 1.9 | 32.0 | 2.4 | 31.1 | 0.9 |
| heart-cleveland | 18.6 | 1.0 | 16.8 | 17.4 | 1.1 | 17.0 | 0.6 | 20.7 | 1.6 | 21.1 | 0.9 |
| hepatitis | 20.1 | 1.6 | 19.1 | 19.5 | 1.7 | 17.8 | 0.7 | 19.0 | 1.3 | 19.7 | 0.9 |
| house-votes-84 | 4.9 | 0.6 | 4.1 | 4.8 | 0.2 | 4.1 | 0.2 | 5.1 | 0.5 | 5.3 | 0.5 |
| hypo | 6.4 | 0.2 | 6.2 | 6.2 | 0.1 | 6.2 | 0.1 | 6.2 | 0.1 | 6.2 | 0.1 |
| ionosphere | 9.7 | 1.3 | 7.4 | 7.5 | 0.5 | 9.2 | 1.2 | 7.6 | 0.6 | 8.3 | 0.5 |
| iris | 4.3 | 1.7 | 2.0 | 3.9 | 0.3 | 4.0 | 0.5 | 3.7 | 0.6 | 3.9 | 1.0 |
| kr-vs-kp | 2.3 | 0.7 | 1.5 | 0.8 | 0.1 | 0.8 | 0.2 | 0.4 | 0.1 | 0.3 | 0.1 |
| labor | 6.1 | 1.5 | 3.5 | 3.2 | 0.8 | 4.2 | 1.0 | 3.2 | 0.8 | 3.2 | 0.8 |
| letter | 18.0 | 0.3 | 17.6 | 12.8 | 0.2 | 10.5 | 0.3 | 5.7 | 0.4 | 4.6 | 0.1 |
| promoters-936 | 5.3 | 0.6 | 4.5 | 4.8 | 0.3 | 4.0 | 0.3 | 4.5 | 0.2 | 4.6 | 0.3 |
| ribosome-bind | 9.3 | 0.4 | 8.9 | 8.5 | 0.3 | 8.4 | 0.4 | 8.1 | 0.2 | 8.2 | 0.3 |
| satellite | 13.0 | 0.3 | 12.6 | 10.9 | 0.2 | 10.6 | 0.3 | 9.9 | 0.2 | 10.0 | 0.3 |
| segmentation | 6.6 | 0.7 | 5.7 | 5.3 | 0.3 | 5.4 | 0.2 | 3.5 | 0.2 | 3.3 | 0.2 |
| sick | 5.9 | 0.5 | 5.2 | 5.7 | 0.2 | 5.7 | 0.1 | 4.7 | 0.2 | 4.5 | 0.3 |
| sonar | 16.6 | 1.5 | 14.9 | 15.9 | 1.2 | 16.8 | 1.1 | 12.9 | 1.5 | 13.0 | 1.5 |
| soybean | 9.2 | 1.1 | 7.0 | 6.7 | 0.5 | 6.9 | 0.4 | 6.7 | 0.5 | 6.3 | 0.6 |
| splice | 4.7 | 0.2 | 4.5 | 4.0 | 0.2 | 3.9 | 0.1 | 4.0 | 0.1 | 4.2 | 0.1 |
| vehicle | 24.9 | 1.2 | 22.9 | 21.2 | 0.8 | 20.7 | 0.6 | 19.1 | 1.0 | 19.7 | 1.0 |

Table 4: Neural network test set error rates and standard deviation values for those error rates for (1) a single neural network classifier; (2) a simple neural network ensemble; (3) a Bagging ensemble; (4) an Arcing ensemble; and (5) and Ada-Boosting ensemble. Also shown (results column 3) is the "best" result produced from all of the single network results run using all of the training data.

| Data Set | Single | | | Bagging | | Arcing | | Boosting | |
|---|---|---|---|---|---|---|---|---|---|
| | Err | SD | Best | Err | SD | Err | SD | Err | SD |
| breast-cancer-w | 5.0 | 0.7 | 4.0 | 3.7 | 0.5 | 3.5 | 0.6 | 3.5 | 0.3 |
| credit-a | 14.9 | 0.8 | 14.2 | 13.4 | 0.5 | 14.0 | 0.9 | 13.7 | 0.5 |
| credit-g | 29.6 | 1.0 | 28.7 | 25.2 | 0.7 | 25.9 | 1.0 | 26.7 | 0.4 |
| diabetes | 27.8 | 1.0 | 26.7 | 24.4 | 0.8 | 26.0 | 0.6 | 25.7 | 0.6 |
| glass | 31.3 | 2.1 | 28.5 | 25.8 | 0.7 | 25.5 | 1.4 | 23.3 | 1.3 |
| heart-cleveland | 24.3 | 1.3 | 22.7 | 19.5 | 0.7 | 21.5 | 1.6 | 20.8 | 1.0 |
| hepatitis | 21.2 | 1.2 | 20.0 | 17.3 | 2.0 | 16.9 | 1.1 | 17.2 | 1.3 |
| house-votes-84 | 3.6 | 0.3 | 3.2 | 3.6 | 0.2 | 5.0 | 1.1 | 4.8 | 1.0 |
| hypo | 0.5 | 0.1 | 0.4 | 0.4 | 0.0 | 0.4 | 0.1 | 0.4 | 0.0 |
| ionosphere | 8.1 | 0.7 | 7.1 | 6.4 | 0.6 | 6.0 | 0.5 | 6.1 | 0.5 |
| iris | 5.2 | 0.7 | 5.3 | 4.9 | 0.8 | 5.1 | 0.6 | 5.6 | 1.1 |
| kr-vs-kp | 0.6 | 0.1 | 0.5 | 0.6 | 0.1 | 0.3 | 0.1 | 0.4 | 0.0 |
| labor | 16.5 | 3.4 | 12.7 | 13.7 | 0.8 | 13.0 | 2.9 | 11.6 | 2.0 |
| letter | 14.0 | 0.8 | 12.2 | 7.0 | 0.1 | 4.1 | 0.1 | 3.9 | 0.1 |
| promoters-936 | 12.8 | 0.4 | 12.5 | 10.6 | 0.6 | 6.8 | 0.5 | 6.4 | 0.3 |
| ribosome-bind | 11.2 | 0.6 | 10.8 | 10.2 | 0.1 | 9.3 | 0.2 | 9.6 | 0.5 |
| satellite | 13.8 | 0.4 | 13.5 | 9.9 | 0.2 | 8.6 | 0.1 | 8.4 | 0.2 |
| segmentation | 3.7 | 0.2 | 3.4 | 3.0 | 0.2 | 1.7 | 0.2 | 1.5 | 0.2 |
| sick | 1.3 | 0.9 | 1.1 | 1.2 | 0.1 | 1.1 | 0.1 | 1.0 | 0.1 |
| sonar | 29.7 | 1.9 | 26.9 | 25.3 | 1.3 | 21.5 | 3.0 | 21.7 | 2.8 |
| soybean | 8.0 | 0.5 | 7.5 | 7.9 | 0.5 | 7.2 | 0.2 | 6.7 | 0.9 |
| splice | 5.9 | 0.3 | 5.7 | 5.4 | 0.2 | 5.1 | 0.1 | 5.3 | 0.2 |
| vehicle | 29.4 | 0.7 | 28.6 | 27.1 | 0.9 | 22.5 | 0.8 | 22.9 | 1.9 |

Table 5: Decision tree test set error rates and standard deviation values for those error rates for (1) a single decision tree classifier; (2) a Bagging ensemble; (3) an Arcing ensemble; and (4) and Ada-Boosting ensemble. Also shown (results column 3) is the "best" result produced from all of the single tree results run using all of the training data.